\documentclass[conference]{IEEEtran}

\usepackage{times}
\usepackage[numbers]{natbib}
\usepackage{multicol}
\usepackage{amsmath}
\usepackage{verbatim}
\usepackage{amssymb}
\usepackage{amsthm}
\usepackage{algorithm}
\usepackage{algorithmic}
\usepackage{graphicx}
\usepackage{subfigure} 
\usepackage{booktabs}
\usepackage[hyphens]{url}
\usepackage{epstopdf}
\usepackage{lipsum}

\PassOptionsToPackage{hyphens}{url}\usepackage[bookmarks=true]{hyperref}

\newcommand{\cD}{{\mathcal{D}}}
\newcommand{\cG}{{\mathcal{G}}}
\newcommand{\cK}{{\mathcal{K}}}

\newcommand{\cN}{{\mathcal{N}}}

\newcommand{\bs}[1]{\boldsymbol{#1}}

\newcommand{\tabref}[1]{Table~\ref{#1}}
\newcommand{\figref}[1]{Fig.~\ref{#1}}
\newcommand{\eqnref}[1]{(\ref{#1})}
\newcommand{\secref}[1]{Section~\ref{#1}}
\newcommand{\algoref}[1]{Algorithm~\ref{#1}}

\newtheorem{algo}{Algorithm}

\newcommand{\ca}[1]{\cite{#1}}

\newcommand{\sdg}{SDGs }
\newcommand{\amr}{AMRs }
\newcommand{\sdgns}{SDGs}
\newcommand{\amrns}{AMRs}
\newcommand{\gdk}{GDK }
\newcommand{\gdkns}{GDK}
\newcommand{\mmd}{MMD }

\newcommand{\mli}{MSI }
\newcommand{\mlins}{MSI}
\newcommand{\sli}{SLI }
\newcommand{\slins}{SLI}

\allowdisplaybreaks

\begin{document}
\onecolumn

% paper title
\title{Extracting Biomolecular Interactions Using Semantic Parsing of Biomedical Text}

% You will get a Paper-ID when submitting a pdf file to the conference system
\author{Sahil Garg, Aram Galstyan, Ulf Hermjakob, and Daniel Marcu\\
USC Information Sciences Institute\\
Marina del Rey, CA 90292\\
\{sahil, galstyan, ulf, marcu\}@isi.edu
}

\maketitle

\begin{abstract}
We advance the state of the art in biomolecular interaction extraction with three contributions: (i)  We show that deep, Abstract Meaning Representations (AMR) significantly improve the accuracy of a biomolecular interaction extraction system when compared to a baseline that relies solely on surface- and syntax-based features; (ii) In contrast with previous approaches that infer relations on a sentence-by-sentence basis, we expand our framework to enable consistent predictions over sets of sentences (documents); (iii) We further modify and expand a graph kernel learning framework to enable concurrent exploitation of automatically induced AMR (semantic) and dependency structure (syntactic) representations. Our experiments show that our approach yields interaction extraction systems that are more robust in environments where there is a significant mismatch between training and test conditions. 
\end{abstract}

\IEEEpeerreviewmaketitle

\section{Introduction}
\footnote{This work has been published previously~\cite{garg2016extracting}.}
Recent advances in genomics and proteomics have significantly accelerated the rate of uncovering and accumulating new biomedical knowledge. Most of this knowledge is available only via scientific publications, which necessitates the development of automated and semi-automated tools for extracting useful biomedical information from unstructured text. In particular, there has been a significant body of research on identifying biological entities (proteins, genes, chemical compounds) and interactions between those entities from bio-medical papers~\ca{krallinger2008overview,hakenberg2008gene,tikk2010comprehensive,bunescu2005comparative}. Despite the recent progress, current methods for biomedical knowledge extraction suffer from a number of important shortcomings. First of all, existing methods rely heavily on shallow analysis techniques that severely limit their scope. For instance, most existing approaches focus on whether there is an interaction between a pair of proteins while ignoring the interaction types~\ca{airola2008all,mooney2005subsequence}, whereas other more advanced approaches cover only a small subset of all possible interaction types~\ca{hunter2008opendmap,mcdonald2005simple,demir2010biopax}. Second, most existing methods focus on single-sentence extraction, which makes them very susceptible to noise. And finally, owing to the enormous diversity of research topics in biomedical literature and the high cost of data annotation, there is often significant  mismatch between training and testing corpora, which reflects poorly on generalization ability of existing methods~\ca{tikk2010comprehensive}.
	    
In this paper, we present a novel algorithm for extracting biomolecular interactions from unstructured text that addresses the above challenges. Contrary to the previous works, the extraction task considered here  is less restricted and spans a much more diverse corpus of biomedical articles. These more realistic settings present some important technical problems for which we provide explicit solutions.
	    
Our specific contributions are as follows:
\begin{itemize}
\item We propose a graph-kernel based algorithm for extracting biomolecular interactions from Abstract Meaning Representation, or AMR. To the best of our knowledge, this is the first attempt of using deep semantic parsing for biomedical knowledge extraction task. 
\item
We provide a multi-sentence generalization of the algorithm by defining \emph{Graph Distribution Kernels} (GDK), which enables us to perform document-level extraction. 
\item We suggest a hybrid extraction method that utilizes both AMRs and syntactic parses given by Stanford Dependency Graphs (SDGs). Toward this goal, we develop a linear algebraic formulation  for \emph{learning vector space embedding of edge labels} in \amr and \sdg to define similarity measures between AMRs and SDGs. 
\end{itemize}
    
We conduct an exhaustive empirical evaluation of the proposed extraction system on 45+ research articles on cancer~(approximately 3k sentences), containing approximately 20,000 positive-negative labeled biomolecular interactions\footnote{The code and the data are available at \url{https://github.com/sgarg87/big_mech_isi_gg}}. Our results indicate that the joint extraction method that leverages both \amr and \sdg parses significantly improves the extraction accuracy, and is more robust to mismatch between training and test conditions.
    
\section{Problem Statement}
\label{sec:ps}
Consider the sentence {\em ``As a result, mutant Ras proteins accumulate with elevated GTP-bound proportion"}, which describes a ``binding" interaction between a protein ``Ras" and a small-molecule ``GTP". We want to extract this interaction.

 In our representation, which is motivated by BioPAX~\ca{demir2010biopax}, an {\em interaction} refers to either i) an entity effecting state change of another entity; or ii) an entity binding/dissociating with another entity to form/break a complex while, optionally, also influenced by a third entity. An entity can be of any type existent in a bio pathway, such as protein, complex, enzyme, etc, although here we refer to an entity of all valid types simply as a protein. The change in state of an entity or binding type is simply termed as ``interaction type" in this work. In some cases, entities are capable of changing their state on their own or bind to an instance of its own~(self-interaction). Such special cases are also included. Some examples of interaction types are shown in \tabref{tab:interaction_categories}. 

Below we describe our approach for extracting above-defined interactions from natural language parses of sentences in a research document.

\begin{table}[tp!]
\centering
\fontsize{9}{9}\selectfont
	\renewcommand{\arraystretch}{0.9}% Tighter
	\renewcommand{\tabcolsep}{1.2pt}
	\begin{tabular}{ll}
		\toprule
		\begin{tabular}{l}
		State\\ 
		change
		\end{tabular}
		& 
		\begin{tabular}{l}
			inhibit, 
			phosphorylate,
			signal,
			activate,
			transcript,
			regulate,
			apoptose,
			express,
			\\
			translocate,
			degrade,
			carboxymethylate,
			depalmitoylate,
			acetylate,
			nitrosylate,
			\\
			farnesylate,
			methylate,
			glycosylate,
			hydroxylate,
			ribosylate,
			sumoylate,
			ubiquitinate.
		\end{tabular}		
		\\
		\midrule
		Bind& 
		\begin{tabular}{l}
			bind,
			heterodimerize,
			homodimerize,
			dissociate.
		\end{tabular}		
		\\
		\bottomrule
	\end{tabular}
\caption{
\fontsize{9}{9}\selectfont
Interaction type examples
}
\label{tab:interaction_categories}
\end{table}
	
\section{Extracting Interactions from an AMR}
\label{sec:extract_frm_amrs}
\subsection{AMR Biomedical Corpus}
Abstract Meaning Representation, or AMR, is a semantic annotation of single/multiple sentences~\ca{Banarescu13abstractmeaning}. In contrast to syntactic parses, in AMR, entities are identified, typed and their semantic roles are annotated. AMR maps different syntactic constructs to same conceptual term. For instance, ``binding", ``bound", "bond" correspond to the same concept ``bind-01". Because one AMR representation subsumes multiple syntactic representations, we hypothesize that AMRs have higher utility for extracting biomedical interactions. 
	
We trained an English-to-AMR parser~\ca{pust2015using} on
two manually annotated corpora: i) a corpus
of \emph{17k general domain sentences} including newswire and
web text as published by the Linguistic Data Consortium; and
ii) \emph{3.4k systems biology sentences},
including in-domain PubMedCentral papers and the BEL BioCreative corpus. As part of building the bio-specific AMR corpus, we extended
the PropBank-based framesets used in AMR by 45 bio-specific
frames such as ``phosphorylate-01", ``immunoblot-01" and extended the list of AMR standard named
entities by 15 types such as ``enzyme", ``pathway". It is important to note that these extensions are not specific to biomolecular interactions, and cover more general cancer biology concepts.
	
\subsection{Extracting Interactions}
\begin{figure}[tp]
\centering
\includegraphics[height=65mm, width=0.65\columnwidth]{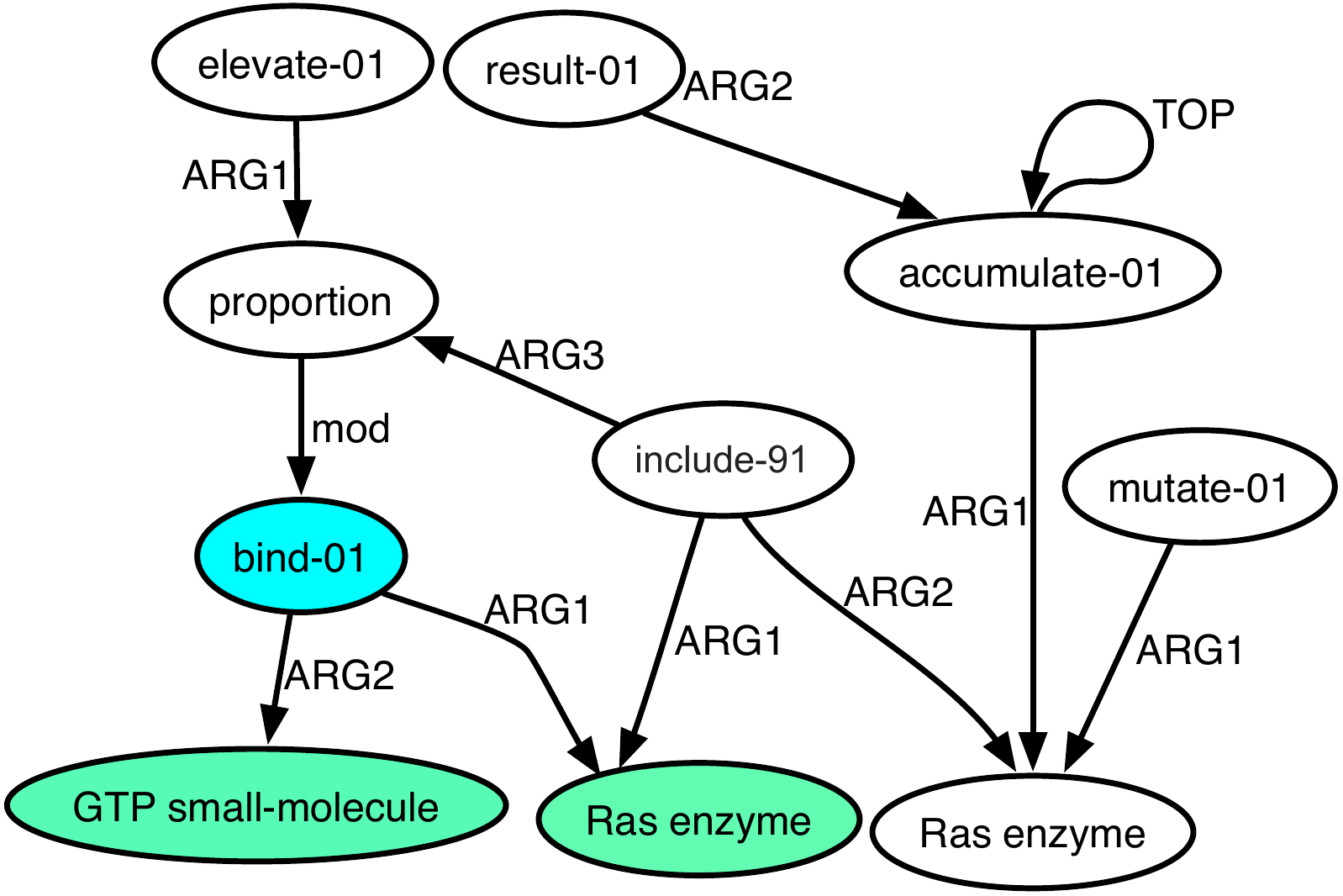}
\caption{
%\fontsize{9}{9}\selectfont
AMR of text ``\textit{As a result, mutant Ras proteins accumulate with elevated GTP-bound proportion.}"; interaction ``\emph{Ras binds to GTP}" is extracted from the colored sub-graph.}
\label{fig:amr_sdg_parse}
\end{figure}
		
 \figref{fig:amr_sdg_parse} depicts a manual AMR annotation of a sentence, which has two highlighted entity nodes with labels ``RAS" and ``GTP". These nodes also have entity type annotations, ``enzyme" and ``small-molecule" respectively; the concept node with a node label ``bind-01" corresponds to an interaction type ``binding"~(from the ``GTP-bound" in the text). The interaction ``RAS-binds-GTP" is extracted from the highlighted subgraph under the ``bind" node. In the subgraph, relationship between the interaction node ``bind-01" and the entity nodes, ``Ras" and ``GTP", is defined through two edges with edge labels ``ARG1" and ``ARG2" respectively. Additionally, in the subgraph, we assign roles ``interaction-type", ``protein", ``protein" to the nodes ``bind-01", ``Ras", ``GTP" respectively~(roles presented with different colors in the subgraph).
	
Given an AMR graph, as in \figref{fig:amr_sdg_parse}, we first identify potential entity nodes~(proteins, molecules, etc) and interaction nodes~(bind, activate, etc). Next, we consider all permutations to generate a set of potential interactions according to the format defined above. For each candidate interaction, we extract the corresponding shortest path subgraph. We then project the subgraph to a tree structure~\footnote{This can be done via so called inverse edge labels; see \cite[section 3]{Banarescu13abstractmeaning}.} with the interaction node as root and also possibly the protein nodes~(entities involved in the interaction) as leaves.

Our training set consists of tuples $\{G^a_i,I_i, l_i\}_{i=1}^n$, where $G^a_i$ is an AMR subgraph constructed such that it can represent an extracted candidate interaction $I_i$ with interaction node as root and proteins nodes as leaves typically; and $l=\{0,1\}$ is a binary label indicating whether this subgraph contains $I_i$ or not.
Given a training set, and a new sample AMR subgraph $G^a_*$ for interaction $I_*$, we would like to infer whether $I_*$ is valid or not.
We address this problem by developing a graph-kernel based approach. 

\subsection{Semantic Embedding Based Graph Kernel}
\label{sec:gk_amr}
We propose an extension of the \emph{contiguous subtree kernel}~\ca{zelenko2003kernel,culotta2004dependency} for mapping the extracted subgraphs~(tree structure) to an implicit feature space. Originally, this kernel uses an identity function on two node labels when calculating the similarity between those two nodes. We instead propose to use vector space embedding of the node labels~\ca{clark2014vector,mikolov2013efficient}, and then define a sparse RBF kernel on the node label vectors. Similar extensions of convolution kernels have been been suggested previously\ca{mehdad2010syntactic,srivastava2013walk}.

Consider two graphs $G_i$ and $G_j$ rooted at nodes $G_i.r$ and $G_j.r$, respectively, and let $G_i.c$ and $G_j.c$ be the children nodes of the corresponding root nodes. Then the kernel between $G_i$ and $G_j$ is defined as follows: 
\[
K(G_i, G_j) 
= 
\begin{cases}
0 & \text{if } k(i,j)=0\\  

k(i,j) 
+
K_c(G_i.c, G_j.c)
&
\text{otherwise}\\
\end{cases}
\ ,
\]
where $k(i,j)\equiv k(G_i.r, G_j.r)$ is the similarity between the root nodes, whereas $K_c(G_i.c, G_j.c)$ is the recursive part of the kernel that measures the similarity of the children subgraphs. Furthermore, the similarity between root nodes $x$ and $y$ is defined as follows:
\begin{align}
\begin{split}
&k(x,y) = k_w(x,y)^2(k_w(x,y)^2+k_e(x,y)+k_r(x,y))
\\
&k_w(x, y)
=
\exp
\left(
		\frac{
		(
		\bs{w}_x^T
		 \bs{w}_y
		 -
		 1
		 )}{
		 \beta}
\right)
\left(
		\frac{
		(\bs{w}_x^T\bs{w}_y-\alpha)
		}{
		(1-\alpha)
		}
\right)_{+}
\\
&
k_e(x,y)
=
\mathbb{I}(e_x = e_y)
,\ 
k_r(x,y)
=
\mathbb{I}(r_x = r_y) \ . \
\end{split}
\label{eqn:k_nodes}
\end{align}
\noindent 
Here $(\cdot)_+$ denotes the positive part; $\mathbb{I}(\cdot)$ is the indicator function; $\bs{w}_x, \bs{w}_y$ are unit vector embeddings of node labels~\footnote{Learned using word2vec software~\ca{mikolov2013efficient} on over one million PubMed articles.}; $e_x, e_y$ represent edge labels~(label of an edge from a node's parent to it is the node's edge label); $r_x, r_y$ are roles of nodes~(such as protein, catalyst, concept, interaction-type); $\alpha$ is a threshold parameter on the cosine similarity~($\bs{w}_x^T\bs{w}_y$) to control  sparsity~\ca{gneiting2002compactly}; and $\beta$ is the bandwidth.
	
The recursive part of the kernel, $K_c$, is defined as follows:
\begin{align*}
\begin{split}
&
K_c(G_i.c, G_j.c)
=
\sum_{\bs{i}, \bs{j}: l(\bs{i})=l(\bs{j})}
\lambda^{l(\bs{i})}
\sum_{s=1,\cdots,l(\bs{i})}
K(G_i[\bs{i}[s]], G_j[\bs{j}[s]])
\prod_{s=1,\cdots,l(\bs{i})}
k(G_i[\bs{i}[s]].r, G_j[\bs{j}[s]].r),
\end{split}
\label{eqn:Kc}
\end{align*}
\noindent where $\bs{i}, \bs{j}$ are contiguous children subsequences under the respective root nodes $G_i.r, G_j.r$; $\lambda \in (0,1)$ is a tuning parameter; and $l(\bs{i})$ is the length of sequence $\bs{i}=i_1, \cdots, i_l$; $G_i[\bs{i}[s]]$ is a sub-tree rooted at $\bs{i}[s]$ index child node of $G_i.r$. Here, we propose to sort children of a node based on the corresponding edge labels. This helps in distinguishing between two mirror image trees.
	
This extension is a valid kernel function~(\citeauthor[Theorem 3, p. 1090]{zelenko2003kernel}). Next, we generalize the dynamic programming approach of~\citeauthor{zelenko2003kernel} for efficient calculation of this extended kernel.
\subsubsection{Dynamic programming for computing convolution graph kernel}
In the convolution kernel presented above, the main computational cost is due to comparison of children sub-sequences. Since different children sub-sequences of a given root node partially overlap with each other, one can use dynamic programming to avoid redundant computations, thus reducing the cost. Toward this goal, we use the following decomposition of the kernel $K_c:$
\begin{align*}
&K_c(G_i.c, G_j.c)
=
\sum_{p,q}
C_{p,q} \ ,
\end{align*}
\noindent where $C_{p,q}$ refers to the similarity between  sub-sequences starting at indices p, q respectively in $G_i.c$ and $G_j.c$. 
	
To calculate $C_{p,q}$ via dynamic programming, let us introduce
\begin{align*}
&L_{p,q}
=
\max_l
\biggl (
	\prod_{s=0}^l
	k(G_i[\bs{i}[p+s]].r, G_j[\bs{j}[q+s]].r)
	\not 
	=
	0	
\biggr ).
\end{align*}
Furthermore, let us denote $k_{p,q}=k(G_i[\bs{i}[p]].r, G_j[\bs{j}[q]].r)$, and $K_{p,q}=K(G_i[\bs{i}[p]], G_j[\bs{j}[q]])$. We then evaluate $C_{p,q}$ in a recursive manner using the following equations.
\begin{align}
\begin{split}
L_{p,q}
=
\begin{cases}
0 & \text{if } k_{p,q} = 0 \\
L_{p+1,q+1}+1 & \text{otherwise}\\
\end{cases}
\end{split}
\end{align}

\begin{align}
\begin{split}
C_{p,q}
=
\begin{cases}
0 
&
\text{if } k_{p,q} = 0 \\
\frac{\lambda(1-\lambda^{L(p,q)})}{1-\lambda}
K_{p,q} k_{p,q} + \lambda C_{p+1, q+1}
& \text{otherwise} \\
\end{cases}
\end{split}
\end{align}
	
\begin{align}
\begin{split}
L_{m+1, n+1} = 0, L_{m+1, n} = 0, L_{m, n+1}=0\\
C_{m+1, n+1} = 0, C_{m+1, n} = 0, C_{m, n+1}=0\ ,
\end{split}
\end{align}
\noindent where $m, n$ are number of children under the root nodes $G_i.r$ and $G_j.r$ respectively.
		
Note that for  graphs with cycles, the above dynamic program can be transformed into a linear program.

There are a couple of practical considerations during the kernel computations. First of all, the kernel depends on two tunable parameters $\lambda$ and $\alpha$. Intuitively, decreasing $\lambda$ discounts the contributions of longer child sub-sequences. The parameter $\alpha$, on the other hand, controls the tradeoff between  computational cost and accuracy.  Based on some prior tuning we found that our results are not very sensitive to the parameters. In the experiments below we set $\lambda=0.99$ and $\alpha=0.4$. Also, consistent with previous studies, we normalize the graph kernel (e.g., kernel similarity $K(G_i, G_j)$ is divided by the normalization term $\sqrt{K(G_i, G_i)K(G_j, G_j)}$) to increase accuracy.	
	
\section{Graph Distribution Kernel- GDK}
\label{sec:gdk}
Often an interaction is mentioned more than once in the same research paper, which justifies a document-level extraction, where one combines evidence from multiple sentences. The prevailing approach to document-level extraction is to first perform inference at sentence level, and then combine those inferences using some type of an aggregation function for a final document-level inference~\ca{skounakis2003evidence,bunescu2006integrating}. For instance, in \ca{bunescu2006integrating}, the inference with the maximum score is chosen. We term this baseline approach as ``Maximum Score Inference", or MSI. Here we advocate a different approach, where one uses the evidences from multiple sentences jointly, for a collective inference. 
		
Let us assume an interaction $I_m$ is supported by  $k_m$ sentences, and  let $\{ G_{m1}, \cdots, G_{m{k_m}} \}$ be the set of relevant AMR subgraphs extracted from those sentences. We can view the elements of this set as samples from some distribution over the graphs, which, with a slight abuse of notation, we denote as  $\cG_m$. Consider now interactions $I_1, \cdots, I_p$, and let $\cG_1, \cdots, \cG_p$ be graph distributions representing  these interactions. 
	
The graph distribution kernel~(\gdkns), $\cK(\cG_i, \cG_j)$, for a pair $\cG_i, \cG_j$ is defined as follows: 
\begin{align*}
\begin{split}
&
\cK(\cG_i, \cG_j)
=
\exp(
-
\mathcal{D}_{mm}(
	\cG_i, \cG_j)
);
\\
& 
\mathcal{D}_{mm}(\cG_i, \cG_j)
=
\sum_{r,s=1}^{k_i}
\frac{
K(G_{ir},G_{is})
}{k_i^2}
+ 
\sum_{r,s=1}^{k_j}
\frac{
K(G_{jr},G_{js})
}{k_j^2}
-
2
\sum_{r,s=1}^{k_i,k_j}
\frac{
K(G_{ir},G_{js})
}{k_ik_j}
\end{split}
\end{align*}
Here $\mathcal{D}_{mm}$ is the \emph{Maximum Mean Discrepancy} (MMD), a valid $l_2$ norm, between a pair of distributions $\cG_i, \cG_j$~\ca{gretton2012kernel}; $K(.,.)$ is the graph kernel defined in \secref{sec:gk_amr}~(though, not restricted to this specific kernel). As the term suggests, \emph{maximum mean discrepancy represents the discrepancy between the mean of graph kernel features~(features implied by kernels) in samples of distributions $\cG_i$ and $\cG_j$. Now, since $\mathcal{D}_{mm}$ is the $l_2$ norm on the mean feature vectors, $\cK(\mathcal{G}_p, \mathcal{G}_q)$ is a valid kernel function.}
		
We note that MMD metric has attracted a considerable attention in the machine learning community recently~\ca{gretton2012kernel,kim2013maximum,pan2008transfer,borgwardt2006integrating}. For our purpose, we prefer using this divergence metric over others~(such as KL-D divergence) for the following reasons: i) $\mathcal{D}_{mm}(.,.)$ is a ``kernel trick" based formulation, nicely fitting with our settings since we do not have explicit features representation of the graphs but only kernel density on the graph samples. Same is true for KL-D estimation with kernel density method. ii) Empirical estimate of $\mathcal{D}_{mm}(.,.)$ is a valid $l_2$ norm distance. Therefore, it is straightforward to derive the graph distribution kernel $\cK(\cG_i, \cG_j)$ from $\mathcal{D}_{mm}(\cG_i, \cG_j)$ using a function such as RBF. This is not true for divergence metrics such as KL-D, Renyi~\ca{sutherland2012kernels}; iii) It is suitable for compactly supported distributions~(small number of samples) whereas methods, such as k-nearest neighbor estimation of KL-D, are not suitable if the number of samples in a distribution is too small~\ca{wang2009divergence}; iv) We have seen the most consistent results in our extraction experiments using this metric as opposed to the others.
	
For the above mentioned reasons, here we focus on MMD as our primary metric for computing similarities between graph distributions. The proposed GDK framework, however, is very general and not limited to a specific metric. Next, we briefly describe two other metrics that can be used with GDK.

\paragraph{GDK with Kullback-Leibler divergence}
While MMD represents maximum discrepancy between the mean features of two distributions, the Kullback-Leibler divergence (KL-D) is a more comprehensive (and fundamental) measure of distance between two distributions\footnote{Recall that the KL divergence between distributions $p$ and $q$ is defined as $\cD_{KL}(p||q)= {\mathbb E_{p(x)} } [\log\frac{p(x)}{q(x)}]$}. For defining kernel $\cK_{KL}$ in terms of KL-D, however, we have two challenges. First of all, KL-D is not a symmetric function. This problem can be addressed by using a symmetric version of the distance in the RBF kernel,
\begin{align*}
\cK_{KL}(\cG_i, \cG_j)
=
\exp(-[\cD_{KL}(\cG_i|| \cG_j)+\cD_{KL}(\cG_j|| \cG_i)])
\end{align*}
where $\cD_{KL}(\cG_i|| \cG_j)$ is the KL distance of the distribution $\cG_i$ w.r.t. the distribution $\cG_j$. And second, even the symmetric combination of the divergences is not a valid Euclidian distance. Hence, $\cK_{KL}$ is not guaranteed to be a positive semi-definite function. This issue can be dealt in a practical manner as nicely discussed in \ca{sutherland2012kernels}. Namely, having computed the Gram matrix using $\cK_{KL}$, we can project it onto a positive semi-definite one by using linear algebraic techniques, e.g., by discarding negative eigenvalues from the spectrum. 
	 
Since we do not know the true divergence, we approximate it with its empirical estimate from the data,  $\cD_{KL}(\cG_i||\cG_j)~\approx~\hat{\cD}_{KL}(\cG_i||\cG_j)$. While there are different approaches for estimating divergences from samples~\ca{wang2009divergence}, here we use kernel density estimator as shown below:
\begin{align*}
\hat{\cD}_{KL}(\cG_i || \cG_j) 
= 
\frac{1}{k_i}
\sum_{r=1}^{k_i} 
\log
\frac{
		\frac{1}{k_i}	
		\sum_{s=1}^{k_i} K(G_{ir}, G_{is})
	}{
		\frac{1}{k_j}
		\sum_{s=1}^{k_j} K(G_{ir}, G_{js})	
}
\end{align*}
\paragraph{GDK with cross kernels}
Another simple way to evaluate similarity between two distributions is to  take the mean of cross-kernel similarities between the corresponding two sample sets:
\begin{align*}
\cK(\cG_i, \cG_j)
=
\sum_{r,s=1}^{k_i,k_j}
\frac{
K(G_{ir},G_{js})
}{k_ik_j}
\end{align*}
Note that this metric looks quite similar to the MMD. As demonstrated in our experiments, however, MMD does better, presumably because it accounts for the mean kernel similarity between samples of the same distribution.
\subsection*{•}
Having defined the graph distribution kernel-\gdkns, $\cK(., .)$, our revised training set consists of tuples $\{\cG_i, I_i, l_i\}_{i=1}^n$ with $G^a_{i1}, \cdots, G^a_{ik_i}$ sample sub-graphs in $\cG_i$. For inferring an interaction $I_*$, we evaluate \gdk between a test distribution $\cG_*$ and the train distributions $\{ \cG_1, \cdots, \cG_n \}$, from their corresponding sample sets. Then, one can apply any ``kernel trick" based classifier.
\section{Cross Representation Similarity}
\label{sec:joint_extraction}
In the previous section, we proposed a novel algorithm for document-level extraction of interactions from \amrns. Looking forward, we will see in our experiments (\secref{sec:exp}) that AMRs yield better extraction accuracy compared to SDGs. This result suggests that using deep semantic features is very useful for the extraction task. On the other hand, the accuracy of semantic (AMR) parsing  is not as good as the accuracy of shallow parsers like \sdgns~\ca{pust2015using,flanigan2014discriminative,wang2015building,andreas2013semantic,chen2014fast}. Thus, one can ask whether the joint use of semantic~(\amrns) and syntactic~(\sdgns) parses can improve extraction accuracy further.
	
There are some intuitive observations that justify the joint approach: i) shallow syntactic parses may be sufficient for correctly extracting a subset of interactions; ii) semantic parsers might make mistakes that are avoidable in syntactic ones. For instance, in machine translation based semantic parsers~\ca{pust2015using,andreas2013semantic}, “hallucinating” phrasal translations may introduce an interaction/protein in a parse that is non-existent in true semantics; iii) over fit of syntactic/semantic parsers can vary from each other in a test corpus depending upon the data used in their independent trainings.
\begin{figure}[tp]
\centering
%\fontsize{8}{8}\selectfont
Abstract Meaning Representation
%\fontsize{7.5}{7.5}\selectfont
\begin{verbatim}
(a / activate-01
   :ARG0 (s / protein :name (n1 / name :op1 "RAS"))
   :ARG1 (s / protein :name (n2 / name :op1 "B-RAF"))
\end{verbatim}
\normalsize
%\fontsize{8}{8}\selectfont
Stanford Typed Dependency
%\fontsize{7.5}{7.5}\selectfont
\begin{verbatim}
nsubj(activates-2, RAS-1)
root(ROOT-0, activates-2)
acomp(activates-2, B-RAF-3)
\end{verbatim}
\caption{
%\fontsize{9}{9}\selectfont
AMR and SDG parses of ``\textit{RAS activates B-RAF.}"
}
\label{fig:amr_sdg_edge_label_ex}
\end{figure}
			
In this setting, in each evidence sentence, a candidate interaction $I_i$ is represented by a tuple $\Sigma_i=\{G^a_i,G^s_i\}$ of sub-graphs $G^a_i$ and $G^s_i$ which are constructed from AMR and SDG parses of a sentence respectively. Our problem is to classify the interaction jointly on features of both sub-graphs. This can be further extended for the use of multiple evidence sentences. We now argue that the graph-kernel framework outlined above can be applied to this setting as well, with some modifications.
	
Let $\Sigma_i$ and $\Sigma_j$ be two sets of points.  To apply the framework above, we need a valid kernel $K(\Sigma_i,\Sigma_j)$ defined on the joint space. One way of defining this kernel would be using similarity measures between AMRs and SDGs separately, and then combining them e.g., via linear combination. However, here we advocate a different approach, where we {\em flatten} the joint representation. Each candidate interaction is represented as a set of two points in the same space. This projection is a valid operation as long as we have a similarity measure between $G^a_i$ and $G^s_i$~(correlation between the two original dimensions). This is rather problematic since AMRs and SDGs have non-overlapping edge labels (although the space of node labels of both representations coincide). To address this issue, for inducing this similarity measure, we next develop our approach for edge-label vector space embedding.
	
Let us understand what we mean by vector space embedding of  edge-labels. In \figref{fig:amr_sdg_edge_label_ex}, we have an AMR and a SDG parse of ``RAS activates B-RAF". ``ARG0" in the AMR and ``nsubj" in SDG are conveying that ``RAS" is a catalyst of the interaction ``activation"; ``ARG1" and ``acomp" are meaning that ``B-RAF" is activated. In this sentence, ``ARG0" and ``nsubj" are playing the same role though their higher dimensional roles, across a diversity set of sentences, would vary. Along these lines, we propose to \emph{embed these high dimensional roles in a vector space}, termed as ``edge label vectors".
	
\subsection{Consistency Equations for Edge Vectors}
\label{sec:evp}
We now describe our unsupervised  algorithm that learns vector space embedding of edge labels. The algorithm works by imposing linear consistency conditions on the word vector embeddings of node labels. While we describe the algorithm using AMRs, it is directly applicable to SDGs as well.
			
\subsubsection{Linear algebraic formulation}
In our formulation, we first learn subspace embedding of edge labels~(edge label matrices) and then transform it into vectors by flattening. Let us see the AMR in \figref{fig:amr_sdg_edge_label_ex} again. We already have word vectors embedding for terms ``activate", ``RAS", ``B-RAF", denoted as $\bs{w}_{activate}$, $\bs{w}_{ras}$, $\bs{w}_{braf}$ respectively; a word vector $\bs{w}_i \in \mathbb{R}^{m \times 1}$. Let embedding for edge labels ``ARG0" and ``ARG1" be $\bs{A}_{arg0}$, $\bs{A}_{arg1}$; $\bs{A}_i \in \mathbb{R}^{m \times m}$.
In this AMR, we define following linear algebraic equations.
\begin{align*}
\begin{split}
&
\bs{w}_{activate}
= 
\bs{A}_{arg0}^T
\bs{w}_{ras}
,
\bs{w}_{activate} 
= 
\bs{A}_{arg1}^T
\bs{w}_{braf} 
\\
&
\bs{A}_{arg0}^T
\bs{w}_{ras}
=
\bs{A}_{arg1}^T
\bs{w}_{braf}
\end{split}
\label{eqn:ex_linear_consistencies}
\end{align*}
The edge label matrices $\bs{A}_{arg0}^T$, $\bs{A}_{arg1}^T$ are linear transformations on the word vectors $\bs{w}_{ras}$, $\bs{w}_{braf}$, establishing linear consistencies between the word vectors along the edges. One can define such a set of equations in each parent-children nodes sub-graph in a given set of manually annotated \amrns~(and so applies to \sdg independent of \amrns).
Along these lines, for a pair of edge labels $i$, $j$ in \amrns, we have generalized equations as below.
\begin{align*}
&
\bs{Y}_i
= 
\bs{X}_i
\bs{A}_i
,\ 
\bs{Y}_j
= 
\bs{X}_j
\bs{A}_j
,\ 
\bs{Z}_i^{ij}
\bs{A}_i
=
\bs{Z}_j^{ij}
\bs{A}_j
\end{align*}
Here $\bs{A}_i, \bs{A}_{j}$ are edge labels matrices. Considering $n_i$  occurrences of edge labels $i$, we correspondingly have word  vectors from the $n_i$ child node labels stacked as rows in matrix $\bs{X}_i \in \mathbb{R}^{n_i \times m}$; and $\bs{Y}_i \in \mathbb{R}^{n_i \times m}$ from the parent node labels. There would be a subset of instances, $n_{ij} <= n_i, n_j$ where edge labels $i$ and $j$ has same parent node~(occurrence of pairwise relationship between $i$ and $j$). This gives $\bs{Z}^{ij}_i \in \mathbb{R}^{n_{ij} \times m}$  and $\bs{Z}^{ij}_j \in \mathbb{R}^{n_{ij} \times m}$, subsets of word vectors in $\bs{X}_{i}$ and $\bs{X}_j$ respectively~(along rows). Along these lines, neighborhood of edge label $i$ is defined to be: $\cN(i): j \in  \cN(i)\ s.t.\ n_{ij} > 0$. From the above pairwise linear consistencies, we derive linear dependencies of an $\bs{A}_i$ with its neighbors $\bs{A}_j: j \in  \cN(i)$, while also applying least square approximation.
\begin{align*}
\begin{split}
&
\bs{X}_i^T \bs{Y}_i
+
\sum_{j \in \cN(i)}
{\bs{Z}_i^{ij}}^T
\bs{Z}_{j}^{ij}
\bs{A}_{j}
= 
(
\bs{X}_i^T \bs{X}_i
+
\sum_{j \in \cN(i)}
{\bs{Z}_i^{ij}}^T 
\bs{Z}_i^{ij}
)
\bs{A}_i
\end{split}
\end{align*}
Exploiting the block structure in the linear program, we propose an algorithm that is a variant of ``Gauss-Seidel" method~\ca{demmel1997applied,niethammer1984convergence}.
\begin{algo}
(a) Initialize:
\[
\bs{A}_i^{(0)}
= 
(\bs{X}_i^T \bs{X}_i)^{-1}
\bs{X}_i^T \bs{Y}_i.
\]
(b) Iteratively update $\bs{A}_i^{(t+1)}$ until convergence: 
\begin{align*}
\begin{split}
&
\bs{A}_i^{(t+1)}
= 
\left[
\bs{X}_i^T \bs{X}_i
+
\sum_{j \in \cN(i)}
{\bs{Z}_i^{ij}}^T 
\bs{Z}_i^{ij}
\right]^{-1}
\left[
\bs{X}_i^T \bs{Y}_i
+
\sum_{j \in \cN(i)}
{\bs{Z}_i^{ij}}^T
\bs{Z}_{j}^{ij}
\bs{A}_{j}^{(t)}
\right]
\end{split}
\end{align*}
(c) Set the inverse edge label matrices: 
\[
\bs{A}_{i_{inv}} = \bs{A}_i^{-1}.
\]
\label{algo:elm}
\end{algo}
Theorem 6.2 in \ca{demmel1997applied}[p. 287, chapter 6] states that the Gauss-Seidel method converges if the linear transformation matrix in a linear program is strictly row diagonal dominant~\ca{niethammer1984convergence}. In our formulation, diagonal blocks dominate the non-diagonal ones row-wise. Thus, \algoref{algo:elm} should converge to an optimum.
		
Using \algoref{algo:elm}, we learned edge label matrices in \amr and \sdg independently on corresponding \amr and \sdg annotations from 2500 bio-sentences~(high accuracy auto-parse for \sdgns). Convergence was fast for both \amr and \sdg~(log error drops from 10.14 to 10.02 for \amrns, and from 30 to approx. 10 for \sdgns).

Next, we flatten an edge label matrix $\bs{A}_i \in \mathbb{R}^{m \times m}$ to a corresponding edge label vector~\footnote{alternatives for kernel directly on the matrices instead of the flattening can be more accurate, that we plan to explore in the future} $\bs{e}_i \in \mathbb{R}^{m^2 \times 1}$, and then redefine $k_e(x,y)$ in \eqnref{eqn:k_nodes} using the sparse RBF kernel.
\[
k_e(x, y)
=
\exp
\left(
		\frac{
		\bs{e}_x^T
		 \bs{e}_y
		 -
		 1
		 }{
		 \beta
		 }
\right)
\left(
		\frac{
			\bs{e}_x^T\bs{e}_y
			-
			\alpha
		}{
		1
		-
		\alpha
		}
\right)_{+}
\] 
This redefinition enables to define kernel similarity between \amr and \sdgns. One can either use our original formulation where a single AMR/SDG sub-graph is classified using training sub-graphs from both \amr and \sdgns, and then the inference with maximum score-\mli~\ca{bunescu2006integrating} is chosen. Another option, preferable, is to consider the set $\{ G^a_i, G^s_i \}$ as samples of a graph distribution $\cG_i$ representing an interaction $I_i$. Generalizing it further, $\cG_i$ has samples set $\{ G^a_{i1}, \cdots, G^a_{ik_{i}^a}, G^s_{i1}, \cdots, G^s_{ik_{i}^s} \}$, containing $k_i^a$, $k_i^s$ number of sub-graphs in \amr and \sdg respectively from multiple sentences in a document, all for classifying $I_i$. With this graph distribution representation, we can apply our \gdk from \secref{sec:gdk} and then infer using a ``kernel trick" based classifier. This final formulation gives the best results in our experiments discussed next.
\section{Experimental Evaluation}
\label{sec:exp}
We evaluated the proposed algorithm on two data sets.
		
\subsection{Data Sets}
\subsubsection{PubMed45}
This dataset has 400 manual and 3k auto parses of \amrns~(and 3.4k auto parses of \sdgns)\footnote{not the same 2.5k sentences used in learning edge label vectors}; \amr auto-parses are from 45 PubMed articles on cancer. From the 3.4k \amrns, we extract 25k subgraphs representing 20k interactions~(valid/invalid); same applies to \sdgns. This is our primary data for the evaluation.
    
We found that for both AMR and SGD based methods, a  part of the extraction error can be attributed to poor recognition of named entities. To minimize this effect, and to isolate errors that are specific to  the extraction methods themselves, we follow the footsteps of the previous studies, and take a filtered subset of the interactions~(approx. 10k out of 20k). We refer to this data subset as ``PubMed45" and the super set as ``PubMed45-ERN" (for entity recognition noise).
	
\subsubsection{AIMed}
This is a publicly available dataset\footnote{\url{http://corpora.informatik.hu-berlin.de}}, which contains about 2000 sentences from 225 abstracts. In contrast to PubMed45, this dataset is very limited as it describes only whether a given pair of proteins interact or not, without specifying the interaction type. Nevertheless, we find it useful to include this dataset in our evaluation since it enables us to compare our results with other reported methods.
	
\subsection{Evaluation settings}
In a typical evaluation scenario, validation is performed by random sub-sampling of labeled interactions~(at sentence level) for a test subset, and using the rest as a training set. This sentence-level validation approach is not always appropriate for extracting protein interactions~\ca{tikk2010comprehensive}, since interactions from a single/multiple sentences in a document can be correlated. Such correlations can lead to information leakage between training and test sets (artificial match, not encountered in real settings). For instance, in \ca{mooney2005subsequence}, the reported F1 score from the random validation in the AIMed data is approx. 0.5. Our algorithm, even using \sdgns, gives 0.66 F1 score in those settings. However, the performance drops significantly when an independent test document is processed. Therefore, for a realistic evaluation, we divide data sets at documents level into approx. 10 subsets such that there is minimal match between a subset, chosen as test set, and the rest of sub sets used for training a kernel classifier. In the PubMed45 data sets, the 45 articles are clustered into 11 subsets by clustering PubMed-Ids~(training data also includes gold annotations). In AIMed, abstracts are clustered into 10 subsets on abstract-ids. In each of 25 independent test runs~(5 for AIMed data) on a single test subset, 80\% interactions are randomly sub sampled from the test subset and same percent from the train data.

\begin{table}[tp!]
\centering
%\fontsize{8}{8}\selectfont
%	\renewcommand{\arraystretch}{0.9}% Tighter
%	\renewcommand{\tabcolsep}{1.5pt}
	\begin{tabular}{llll}
		\toprule
		\textbf{Methods}&\textbf{PubMed45-ERN}&\textbf{PubMed45}&\textbf{AIMed}\\
		\toprule
		SDG (\slins)&$0.25\pm0.16$&$0.32\pm0.18$&$0.27\pm0.12$\\
		&$(0.42, 0.29)$&$(0.50, 0.35)$&$(0.54, 0.22)$\\
 		\midrule
		AMR (\slins)&$0.33\pm0.16$&$0.45\pm0.25$&$0.39\pm0.05$\\
		&$(0.33, 0.45)$&$(0.58, 0.43)$&$(0.53, 0.33)$\\
		\toprule
		SDG~(\mlins)&$0.24\pm0.14$&$0.33\pm0.17$&$0.39\pm0.09$\\
		&$(0.39, 0.28)$&$(0.50, 0.34)$&$(0.51, 0.38)$\\
 		\midrule
		AMR~(\mlins)&$0.32\pm0.14$&$0.45\pm0.24$&$0.51\pm0.11$\\
		&$(0.30, 0.45)$&$(0.56, 0.44)$&$(0.49, 0.56)$\\
		\toprule
		SDG~(GDK)&$0.25\pm0.16$&$0.38\pm0.15$&$0.47\pm0.08$\\
		&$(0.33, 0.31)$&$(0.32, 0.61)$&$(0.41, 0.58)$\\
 		\midrule
		AMR~(GDK)&$0.35\pm0.16$&$0.51\pm0.23$&$0.51\pm0.11$\\
		&$(0.31, 0.51)$&$(0.59, 0.49)$&$(0.43, 0.65)$\\
		\toprule
		AMR-SDG~(\mlins)&$0.33\pm0.18$&$0.47\pm0.24$&$\bs{0.55\pm0.09}$\\
		&$(0.29, 0.54)$&$(0.50, 0.53)$&$(0.46, 0.73)$\\
 		\midrule
		AMR-SDG~(GDK)&$\bs{0.38\pm0.16}$&$\bs{0.57\pm0.23}$&$0.52\pm0.09$\\
		&$(0.33, 0.55)$&$(0.63, 0.54)$&$(0.43, 0.67)$\\
		\toprule
		\textbf{Data Statistics}&&&\\
		\toprule
		Positive ratio&$0.07\pm0.04$&$0.19\pm0.14$&$0.37\pm0.11$\\
		Train-Test Div.&$0.014\pm0.019$&$0.041\pm0.069$&$0.005\pm0.002$\\	
		\bottomrule
	\end{tabular}
\caption{
%\fontsize{9}{9}\selectfont
F1 score statistics. ``\slins" is sentence level inference; ``\mlins" refers to maximum score inference at document level; ``\gdkns" denotes Graph distribution kernel based inference at document level. Precision, recall statistics are presented as (mean-precision, mean-recall) tuples.
}
\label{tab:results}
\end{table}
For the classification, we use the LIBSVM implementation of Kernel Support Vector Machines~\ca{chang2011libsvm} with the sklearn python wrapper~\footnote{\url{http://scikit-learn.org/stable/modules/generated/sklearn.svm.SVC.html}}. Specifically, we used settings \{ $probability=True$, $C=1$, $class\_weight=auto$\}. In our data, we have a class ``swap" in addition to the two binary classes~(``valid", ``invalid"). The ``swap" class means that an interaction is invalid as such but swapping of entity roles in the interaction makes it valid. For the analysis purpose however, we focus on F1 scores only for the positive class, i.e. class ``valid".
	
\subsection{Evaluation Results}
We categorize all methods evaluated below as follows: i) \emph{Sentence Level Inference}-SLI~\footnote{Note that even for the sentence level inference, the training/test division is done on  document level.}; ii) document level using \emph{Maximum Score Inference}-\mli~\ca{bunescu2006integrating}; and iii) document-level inference on all the subgraphs using our \emph{Graph Distribution Kernel} (GDK). In each of the categories, \amrns, \sdg are used independently, and then jointly. Edge label vectors are used only when \amr and \sdg are jointly used, referred as ``AMR-SDG".
	
\tabref{tab:results} shows the F1 score statistics for all the experiments. In addition, the mean of precision and recall values are presented as (precision, recall) tuples in the same table. For most of the following discussion, we focus on F1 scores only to keep the exposition simple.
        
Before going into detailed discussion of the results, we make the following two observations. First, we can see that, in all methods~(including our \gdk and baselines), we obtain much better accuracy using \amr compared to \sdgns. This result is remarkable, especially taking into account the fact that the accuracy of semantic parsing is still significantly lower when compared to syntactic parsing. And second, observe that the overall accuracy numbers are considerably lower for the PubMed45-ERN data, compared to the filtered data PubMed45.  
	
Let us focus on document-level extraction using \mlins. We do not see much improvement in numbers compared to \sli for our PubMed45 data. On the other hand, even this simple \mli technique works for the restricted extraction settings in the AIMed data. \mli works for AIMed data probably because there are multiple sub-graph evidences with varying interaction types~(root node in subgraphs), even in a single sentence, all representing same protein-protein pair interaction. This high number of evidences at document level, should give a boost in performance even using \mlins.

Next, we consider document-level extraction using the proposed \gdk method with the MMD metric. Comparing against the baseline \slins, we see a significant improvement for all data sets and in both \amr and \sdg~(although the improvement in PubMed45-ERN is relatively small). The effect of the noise in entity recognition can be a possible reason why \gdk does not work so well in this data compared to the other two data sets. Here, we also see that: a) \gdk method performs better than the document level baseline \mlins; and b) \amr perform better than \sdg with \gdk method also.
	        		
Let us now consider the results of extraction using both \amr and \sdg jointly. Here we evaluate \mli and GDK, both using our edge label vectors. Our primary observation here is that the joint inference using both \amr and \sdg improves the extraction accuracy across all datasets. Furthremore, in both PubMed45 datasets, the proposed \gdk method is a more suitable choice for the joint inference on \amr and \sdgns. As we can see, comparing to \gdk for \amr only, F1 points increment from 0.35 to 0.38 for the PubMed45-ERN data, and from 0.51 to 0.57 for the PubMed45 data. For the AIMed dataset, on the other hand, the best result (F1 score of 0.55) is obtained when one uses the baseline \mli for the joint inference on \amr and \sdgns.  
    
To get more insights, we now consider (mean-precision, mean-recall) tuples shown in the \tabref{tab:results}. The general trend is that the \amr lead to higher recall compared to the \sdgns. In the PubMed45-ERN data set, this increase in the recall is at cost of a drop in the precision values. Since the entity types are noisy in this data set, this drop in the precision numbers is not completely surprising~(note that the F1 scores still increase). With the use of the GDK method in the same data set, however, the precision drop~(\sdg to \amrns) becomes negligible, while the recall still increases significantly. In the data set PubMed45~(the one without noise in the entity types), both the precision and recall are generally higher for the \amr compared to the \sdgns. Again, there is an exception for the GDK approach, for which the recall decreases slightly. However, the corresponding precision almost doubles.
	    		
\begin{figure}[tp]
\centering
\subfigure[PubMed45]{
\includegraphics[height=65mm, width=0.7\columnwidth]{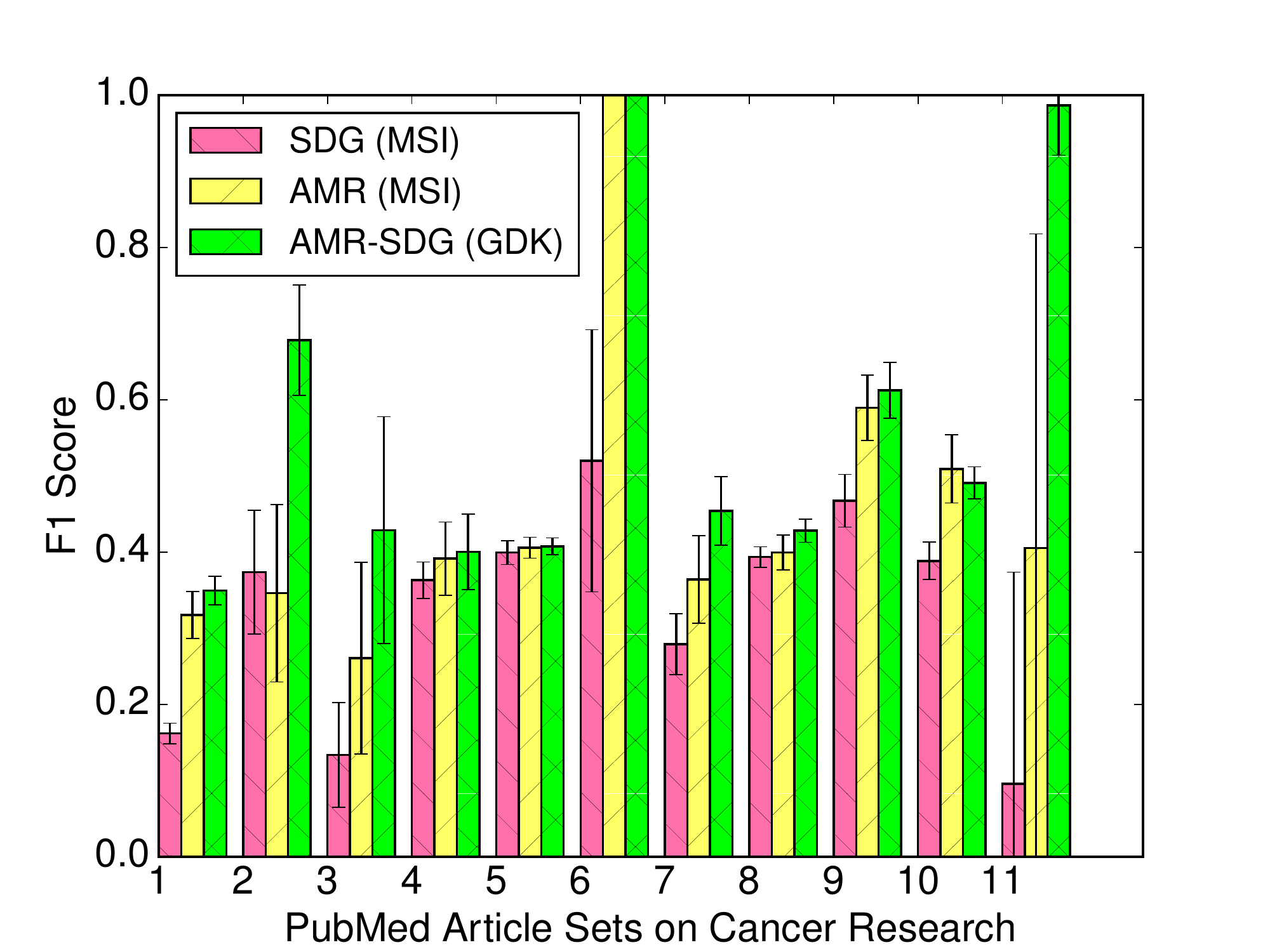}
}
%\hspace{-8.5mm}
\subfigure[AIMed]{
\includegraphics[height=65mm, width=0.7\columnwidth]{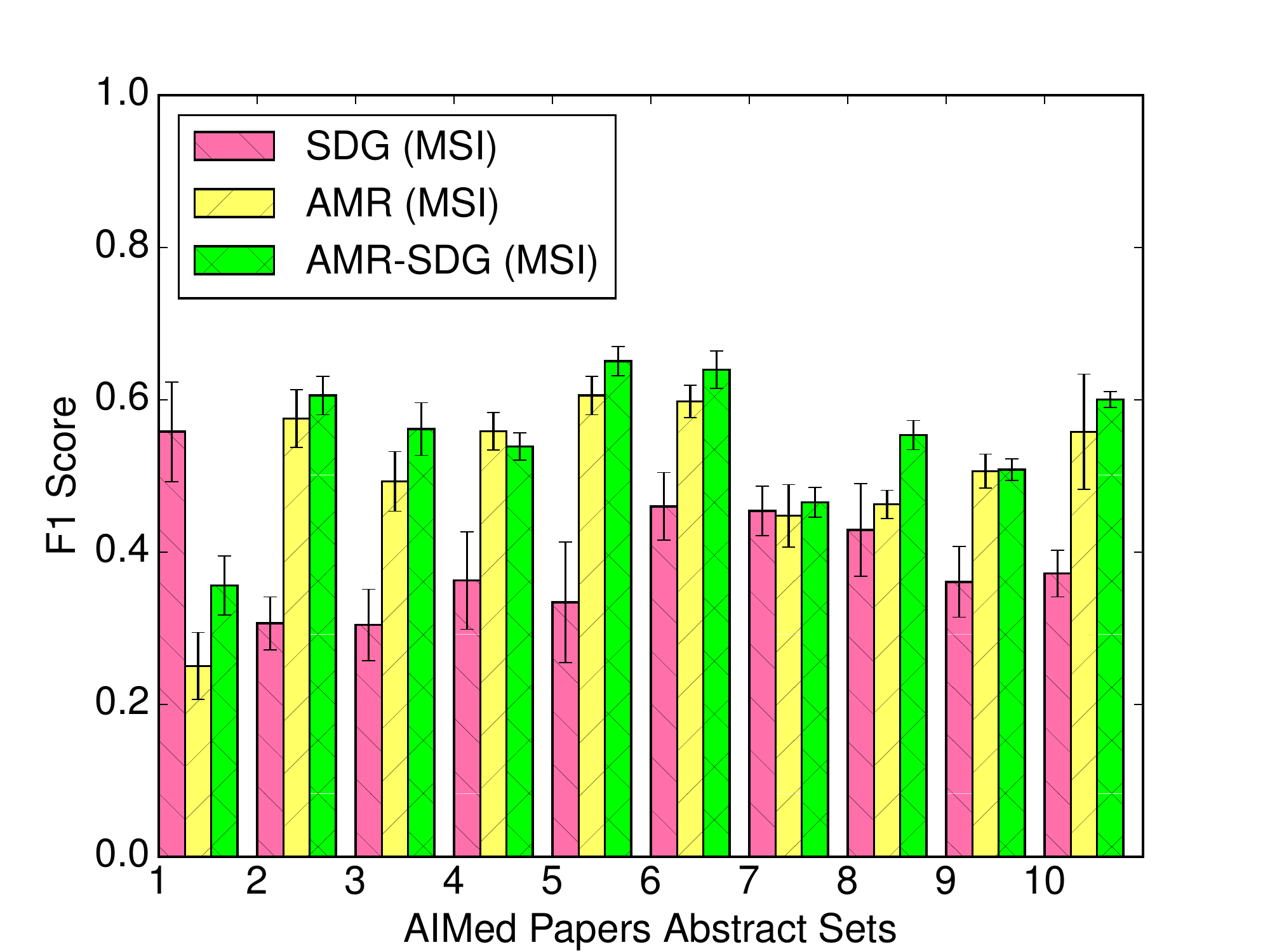}
}
\caption{
%\fontsize{9}{9}\selectfont
Comparison of extraction accuracy~(F1 score)
}
\label{fig:f1_score}
\end{figure}
	
For a more fine-grained comparison between the methods, we plot F1 score for each individual test set in \figref{fig:f1_score}. Here, we compare the baselines, ``AMR (\mlins)", ``SDG (\mlins)" against the ``AMR-SDG (GDK)" in our data~(and, ``AMR-SDG (\mlins)" for ``AIMed"). We see a general trend, across all test subsets, of \amr being more accurate than \sdg and the joint use of two improving even upon \amrns. Though, there are some exceptions where the difference is marginal between the three. In our cross checking, we find that such exceptions are when there is relatively more information leakage between train-test, i.e. less train-test divergence. We use Maximum Mean Discrepancy-\mmd for evaluating this train-test divergence~(originally used for defining \gdk in \secref{sec:gdk}. We find that our \gdk technique is more suitable when $\mmd > 0.01$~(\mmd is normalized metric for a normalized graph kernel).
	
\begin{table}[tp!]
\centering
	\begin{tabular}{llllll}
		\toprule
		\textbf{}&\textbf{MMD}&\textbf{KL-D}&\textbf{CK}\\
		\midrule
 		SDG&$0.25\pm0.16$&$0.21\pm0.17$&$0.26\pm0.13$\\
 		&$(0.33, 0.31)$&$(0.59, 0.21)$&$(0.29, 0.38)$\\
		\midrule
 		AMR&$0.35\pm0.16$&$0.37\pm0.17$&$0.29\pm0.13$\\
 		&$(0.31, 0.51)$&$(0.50, 0.41)$&$(0.28, 0.39)$\\
	\end{tabular}
\caption{
%\fontsize{9}{9}\selectfont
Comparison of F1 scores for different divergence metrics used with GDK. The evaluation is on PubMed45-ERN dataset. ``KL-D" and ``CK" stand for Kullback-Leibler divergence and Cross Kernels, respectively.}
\label{tab:results_other_metrics}
\end{table}
	
The results for the GDK method described above are specific to the MMD metric. We also evaluated GDK using two other metrics~(KL-D and cross kernels), specifically on ``PubMed45-ERN" dataset, as presented in \tabref{tab:results_other_metrics}. Here, as in \tabref{tab:results}, we also present (mean-precision, mean-recall) tuples. We can see that MMD and KL-D metrics, both, perform equally well for AMR whereas MMD does better in case of SDG. CK~(cross kernels), which is a relatively naive approach, also performs reasonably well, although for the AMRs it performs worse compared to  MMD and KL-D. For the precision and recall numbers in the \tabref{tab:results_other_metrics}, we see similar trends as reported in \tabref{tab:results}. We observe that the recall numbers increase for the \amr compared to the \sdg (the metric CK is an exception with negligible increase). Also, comparing KL-D against MMD, we see the former favors (significantly) higher precision, albeit at the expense of lower recall values.

\section{Related Work}
\label{sec:related_work}
There have been different lines of work for extracting protein extractions. Pattern-matching based systems~(either manual or semi-automated) usually yield high precision but low recall~\ca{hunter2008opendmap,krallinger2008overview,hakenberg2008gene,hahn2015domain}. Kernel-based methods based on various convolution kernels have also been developed for the extraction task~\ca{tikk2010comprehensive,airola2008all,mooney2005subsequence}. Some approaches work on string rather than parses~\ca{mooney2005subsequence}. The above mentioned works either rely on text or its shallow parses, none using semantic parsing for the extraction task. Also, most works consider only protein-protein interactions while ignoring interaction types. Some recent works used distant supervision to obtain a large data set of protein-protein pairs for their experiments~\cite{mallory2015large}.
		
Document-level extraction has been explored in the past~\ca{skounakis2003evidence,bunescu2006integrating}. These works classify at sentence level and then combine the inferences whereas we propose to infer jointly on all the sentences at document level.
	
Previously, the idea of linear relational embedding has been explored in \ca{paccanaro2000learning}, where triples of concepts and relation types between those concepts are (jointly) embedded in some latent space. Neural networks have also been employed for joint embedding~\ca{bordes2014semantic}. Here we advocate for a factored embedding where concepts~(node labels) are embedded first using plain text, and then relations~(edge labels) are embedded in a linear sub-space.
	
\section{Conclusion}
\label{sec:conclusions}
In summary, we have developed and validated a method for extracting biomolecular interactions that, for the first time, uses  \emph{deep semantic parses} of biomedical text~(\amrns). We have presented a novel algorithm, which relies on \emph{Graph Distribution Kernels} (GDK) for document-level extraction of interactions from a set of \amr in a document. \gdk can operate on both AMR and SDG parses of sentences jointly. The rationale behind this hybrid approach is that while neither parsing is perfect, their combination can yield superior results. Indeed, our experimental results suggest that the proposed approach outperforms the baselines, especially in the practically relevant scenario when there is a noticeable mismatch between the training and test sets. 

To facilitate the joint approach, we have proposed a novel edge vector space embedding method to assess similarity between different types of parses. We believe this notion of  edge-similarly is quite general and will have applicability for a wider class of problems involving graph kernels. As a future work, we intend to validate this framework on a number of problems such as improving accuracy in \amr parsing with \sdgns.

\section{Acknowledgements}
This work was sponsored by the DARPA Big Mechanism program (W911NF-14-1-0364). It is our pleasure to acknowledge  fruitful discussions with Michael Pust, Kevin Knight, Shuyang Gao, Linhong Zhu, Neal Lawton, Emilio Ferrara, and Greg Ver Steeg. We are also grateful to anonymous reviewers for their valuable feedback.

\bibliographystyle{plainnat}
\bibliography{arxiv}

\begin{thebibliography}{38}
\providecommand{\natexlab}[1]{#1}
\providecommand{\url}[1]{\texttt{#1}}
\expandafter\ifx\csname urlstyle\endcsname\relax
  \providecommand{\doi}[1]{doi: #1}\else
  \providecommand{\doi}{doi: \begingroup \urlstyle{rm}\Url}\fi

\bibitem[Airola et~al.(2008)Airola, Pyysalo, Bj{\"o}rne, Pahikkala, Ginter, and
  Salakoski]{airola2008all}
Antti Airola, Sampo Pyysalo, Jari Bj{\"o}rne, Tapio Pahikkala, Filip Ginter,
  and Tapio Salakoski.
\newblock All-paths graph kernel for protein-protein interaction extraction
  with evaluation of cross-corpus learning.
\newblock \emph{BMC Bioinformatics}, 9:\penalty0 S2, 2008.

\bibitem[Andreas et~al.(2013)Andreas, Vlachos, and Clark]{andreas2013semantic}
Jacob Andreas, Andreas Vlachos, and Stephen Clark.
\newblock Semantic parsing as machine translation.
\newblock In \emph{Proc. of ACL}, pages 47--52, 2013.

\bibitem[Banarescu et~al.(2013)Banarescu, Bonial, Cai, Georgescu, Griffitt,
  Hermjakob, Knight, Koehn, Palmer, and Schneider]{Banarescu13abstractmeaning}
Laura Banarescu, Claire Bonial, Shu Cai, Madalina Georgescu, Kira Griffitt, Ulf
  Hermjakob, Kevin Knight, Philipp Koehn, Martha Palmer, and Nathan Schneider.
\newblock Abstract meaning representation for sembanking.
\newblock In \emph{Proceedings of the 7th Linguistic Annotation Workshop and
  Interoperability with Discourse}, 2013.

\bibitem[Bordes et~al.(2014)Bordes, Glorot, Weston, and
  Bengio]{bordes2014semantic}
Antoine Bordes, Xavier Glorot, Jason Weston, and Yoshua Bengio.
\newblock A semantic matching energy function for learning with
  multi-relational data.
\newblock \emph{Machine Learning}, 94:\penalty0 233--259, 2014.

\bibitem[Borgwardt et~al.(2006)Borgwardt, Gretton, Rasch, Kriegel,
  Sch{\"o}lkopf, and Smola]{borgwardt2006integrating}
Karsten~M Borgwardt, Arthur Gretton, Malte~J Rasch, Hans-Peter Kriegel,
  Bernhard Sch{\"o}lkopf, and Alex~J Smola.
\newblock Integrating structured biological data by kernel maximum mean
  discrepancy.
\newblock \emph{Bioinformatics}, 22:\penalty0 e49--e57, 2006.

\bibitem[Bunescu et~al.(2005)Bunescu, Ge, Kate, Marcotte, Mooney, Ramani, and
  Wong]{bunescu2005comparative}
Razvan Bunescu, Ruifang Ge, Rohit~J Kate, Edward~M Marcotte, Raymond~J Mooney,
  Arun~K Ramani, and Yuk~Wah Wong.
\newblock Comparative experiments on learning information extractors for
  proteins and their interactions.
\newblock \emph{Artificial Intelligence in Medicine}, pages 139--155, 2005.

\bibitem[Bunescu et~al.(2006)Bunescu, Mooney, Ramani, and
  Marcotte]{bunescu2006integrating}
Razvan Bunescu, Raymond Mooney, Arun Ramani, and Edward Marcotte.
\newblock Integrating co-occurrence statistics with information extraction for
  robust retrieval of protein interactions from medline.
\newblock In \emph{Proc. of the Workshop on Linking Natural Language Processing
  and Biology: Towards Deeper Biological Literature Analysis}, 2006.

\bibitem[Chang and Lin(2011)]{chang2011libsvm}
Chih-Chung Chang and Chih-Jen Lin.
\newblock Libsvm: A library for support vector machines.
\newblock \emph{ACM Transactions on Intelligent Systems and Technology (TIST)},
  2:\penalty0 27, 2011.

\bibitem[Chen and Manning(2014)]{chen2014fast}
Danqi Chen and Christopher~D Manning.
\newblock A fast and accurate dependency parser using neural networks.
\newblock In \emph{Proc. of EMNLP}, pages 740--750, 2014.

\bibitem[Clark(2014)]{clark2014vector}
Stephen Clark.
\newblock Vector space models of lexical meaning.
\newblock \emph{Handbook of Contemporary Semantics, 2nd Ed. Blackwell}, 2014.

\bibitem[Culotta and Sorensen(2004)]{culotta2004dependency}
Aron Culotta and Jeffrey Sorensen.
\newblock Dependency tree kernels for relation extraction.
\newblock In \emph{Proc. of ACL}, page 423, 2004.

\bibitem[Demir et~al.(2010)Demir, Cary, Paley, Fukuda, Lemer, Vastrik, Wu,
  D'Eustachio, Schaefer, Luciano, et~al.]{demir2010biopax}
Emek Demir, Michael~P Cary, Suzanne Paley, Ken Fukuda, Christian Lemer, Imre
  Vastrik, Guanming Wu, Peter D'Eustachio, Carl Schaefer, Joanne Luciano,
  et~al.
\newblock The biopax community standard for pathway data sharing.
\newblock \emph{Nature Biotechnology}, 28:\penalty0 935--942, 2010.

\bibitem[Demmel(1997)]{demmel1997applied}
James~W Demmel.
\newblock \emph{Applied numerical linear algebra}.
\newblock Siam, 1997.

\bibitem[Flanigan et~al.(2014)Flanigan, Thomson, Carbonell, Dyer, and
  Smith]{flanigan2014discriminative}
J.~Flanigan, S.~Thomson, J.~Carbonell, C.~Dyer, and N.~A. Smith.
\newblock A discriminative graph-based parser for the abstract meaning
  representation.
\newblock In \emph{Proc. of ACL}, 2014.

\bibitem[Garg et~al.(2016)Garg, Galstyan, Hermjakob, and
  Marcu]{garg2016extracting}
Sahil Garg, Aram Galstyan, Ulf Hermjakob, and Daniel Marcu.
\newblock Extracting biomolecular interactions using semantic parsing of
  biomedical text.
\newblock In \emph{Proc. of AAAI}, 2016.

\bibitem[Gneiting(2002)]{gneiting2002compactly}
Tilmann Gneiting.
\newblock Compactly supported correlation functions.
\newblock \emph{Journal of Multivariate Analysis}, 83:\penalty0 493--508, 2002.

\bibitem[Gretton et~al.(2012)Gretton, Borgwardt, Rasch, Sch{\"o}lkopf, and
  Smola]{gretton2012kernel}
Arthur Gretton, Karsten~M Borgwardt, Malte~J Rasch, Bernhard Sch{\"o}lkopf, and
  Alexander Smola.
\newblock A kernel two-sample test.
\newblock \emph{JMLR}, 13:\penalty0 723--773, 2012.

\bibitem[Hahn and Surdeanu(2015)]{hahn2015domain}
Marco A Valenzuela-Esc{\'a}rcega~Gus Hahn and Powell Thomas Hicks~Mihai
  Surdeanu.
\newblock A domain-independent rule-based framework for event extraction.
\newblock \emph{ACL-IJCNLP 2015}, page 127, 2015.

\bibitem[Hakenberg et~al.(2008)Hakenberg, Plake, Royer, Strobelt, Leser, and
  Schroeder]{hakenberg2008gene}
J{\"o}rg Hakenberg, Conrad Plake, Loic Royer, Hendrik Strobelt, Ulf Leser, and
  Michael Schroeder.
\newblock Gene mention normalization and interaction extraction with context
  models and sentence motifs.
\newblock \emph{Genome Biol}, 9:\penalty0 S14, 2008.

\bibitem[Hunter et~al.(2008)Hunter, Lu, Firby, Baumgartner, Johnson, Ogren, and
  Cohen]{hunter2008opendmap}
Lawrence Hunter, Zhiyong Lu, James Firby, William~A Baumgartner, Helen~L
  Johnson, Philip~V Ogren, and K~Bretonnel Cohen.
\newblock Opendmap: an open source, ontology-driven concept analysis engine,
  with applications to capturing knowledge regarding protein transport, protein
  interactions and cell-type-specific gene expression.
\newblock \emph{BMC Bioinformatics}, 9:\penalty0 78, 2008.

\bibitem[Kim and Pineau(2013)]{kim2013maximum}
Beomjoon Kim and Joelle Pineau.
\newblock Maximum mean discrepancy imitation learning.
\newblock In \emph{Proc. of RSS}, 2013.

\bibitem[Krallinger et~al.(2008)Krallinger, Leitner, Rodriguez-Penagos, and
  Valencia]{krallinger2008overview}
Martin Krallinger, Florian Leitner, Carlos Rodriguez-Penagos, and Alfonso
  Valencia.
\newblock Overview of the protein-protein interaction annotation extraction
  task of biocreative ii.
\newblock \emph{Genome biology}, 9:\penalty0 S4, 2008.

\bibitem[Mallory et~al.(2015)Mallory, Zhang, R{\'e}, and
  Altman]{mallory2015large}
Emily~K Mallory, Ce~Zhang, Christopher R{\'e}, and Russ~B Altman.
\newblock Large-scale extraction of gene interactions from full-text literature
  using deepdive.
\newblock \emph{Bioinformatics}, page btv476, 2015.

\bibitem[McDonald et~al.(2005)McDonald, Pereira, Kulick, Winters, Jin, and
  White]{mcdonald2005simple}
Ryan McDonald, Fernando Pereira, Seth Kulick, Scott Winters, Yang Jin, and Pete
  White.
\newblock Simple algorithms for complex relation extraction with applications
  to biomedical ie.
\newblock In \emph{Proc. of ACL}, 2005.

\bibitem[Mehdad et~al.(2010)Mehdad, Moschitti, and
  Zanzotto]{mehdad2010syntactic}
Yashar Mehdad, Alessandro Moschitti, and Fabio~Massimo Zanzotto.
\newblock Syntactic/semantic structures for textual entailment recognition.
\newblock In \emph{Human Language Technologies: The 2010 Annual Conference of
  the North American Chapter of the Association for Computational Linguistics},
  pages 1020--1028, 2010.

\bibitem[Mikolov et~al.(2013)Mikolov, Sutskever, Chen, Corrado, and
  Dean]{mikolov2013efficient}
Tomas Mikolov, Ilya Sutskever, Kai Chen, Greg~S Corrado, and Jeff Dean.
\newblock Distributed representations of words and phrases and their
  compositionality.
\newblock In \emph{Proc. of NIPS}, 2013.

\bibitem[Mooney and Bunescu(2005)]{mooney2005subsequence}
Raymond~J Mooney and Razvan~C Bunescu.
\newblock Subsequence kernels for relation extraction.
\newblock In \emph{Proc. of NIPS}, pages 171--178, 2005.

\bibitem[Niethammer et~al.(1984)Niethammer, De~Pillis, and
  Varga]{niethammer1984convergence}
W~Niethammer, J~De~Pillis, and RS~Varga.
\newblock Convergence of block iterative methods applied to sparse
  least-squares problems.
\newblock \emph{Linear Algebra and its Applications}, 58:\penalty0 327--341,
  1984.

\bibitem[Paccanaro and Hinton(2000)]{paccanaro2000learning}
Alberto Paccanaro and Geoffrey~E Hinton.
\newblock Learning distributed representations by mapping concepts and
  relations into a linear space.
\newblock In \emph{Proc. of ICML}, pages 711--718, 2000.

\bibitem[Pan et~al.(2008)Pan, Kwok, and Yang]{pan2008transfer}
Sinno~Jialin Pan, James~T Kwok, and Qiang Yang.
\newblock Transfer learning via dimensionality reduction.
\newblock In \emph{Proc. of AAAI}, 2008.

\bibitem[Pust et~al.(2015)Pust, Hermjakob, Knight, Marcu, and
  May]{pust2015using}
Michael Pust, Ulf Hermjakob, Kevin Knight, Daniel Marcu, and Jonathan May.
\newblock Using syntax-based machine translation to parse english into abstract
  meaning representation.
\newblock In \emph{Proc. of EMNLP}, 2015.

\bibitem[Skounakis and Craven(2003)]{skounakis2003evidence}
Marios Skounakis and Mark Craven.
\newblock Evidence combination in biomedical natural-language processing.
\newblock In \emph{Proc. of BIOKDD}, 2003.

\bibitem[Srivastava et~al.(2013)Srivastava, Hovy, and Hovy]{srivastava2013walk}
Shashank Srivastava, Dirk Hovy, and Eduard~H Hovy.
\newblock A walk-based semantically enriched tree kernel over distributed word
  representations.
\newblock In \emph{Proc. of EMNLP}, pages 1411--1416, 2013.

\bibitem[Sutherland et~al.(2012)Sutherland, Xiong, P{\'o}czos, and
  Schneider]{sutherland2012kernels}
Dougal~J Sutherland, Liang Xiong, Barnab{\'a}s P{\'o}czos, and Jeff Schneider.
\newblock Kernels on sample sets via nonparametric divergence estimates.
\newblock \emph{arXiv preprint arXiv:1202.0302}, 2012.

\bibitem[Tikk et~al.(2010)Tikk, Thomas, Palaga, Hakenberg, and
  Leser]{tikk2010comprehensive}
Domonkos Tikk, Philippe Thomas, Peter Palaga, J{\"o}rg Hakenberg, and Ulf
  Leser.
\newblock A comprehensive benchmark of kernel methods to extract
  protein--protein interactions from literature.
\newblock \emph{PLoS Comput Biol}, 2010.

\bibitem[Wang et~al.(2009)Wang, Kulkarni, and Verd{\'u}]{wang2009divergence}
Qing Wang, Sanjeev~R Kulkarni, and Sergio Verd{\'u}.
\newblock Divergence estimation for multidimensional densities
  via-nearest-neighbor distances.
\newblock \emph{IEEE Transactions on Information Theory}, pages 2392--2405,
  2009.

\bibitem[Wang et~al.(2015)Wang, Berant, and Liang]{wang2015building}
Yushi Wang, Jonathan Berant, and Percy Liang.
\newblock Building a semantic parser overnight.
\newblock In \emph{Proc. of ACL}, 2015.

\bibitem[Zelenko et~al.(2003)Zelenko, Aone, and Richardella]{zelenko2003kernel}
Dmitry Zelenko, Chinatsu Aone, and Anthony Richardella.
\newblock Kernel methods for relation extraction.
\newblock \emph{JMLR}, 3:\penalty0 1083--1106, 2003.

\end{thebibliography}

\end{document}